\documentclass[journal, twoside]{IEEEtran}
\let\labelindent\relax
\usepackage{amsmath}
\usepackage{amssymb}
\usepackage{mathtools}
\usepackage{eqlist}
\usepackage{amsfonts}

\usepackage{amsthm}

\usepackage{tabularx}
\usepackage{threeparttable}
\usepackage{booktabs}
\usepackage{makecell}
\usepackage{colortbl}
\usepackage{graphicx}
\usepackage{wrapfig}
\usepackage{subcaption}
\usepackage[lined, ruled, linesnumbered, commentsnumbered, noend]{algorithm2e}
\usepackage[colorlinks=true, allcolors=blue]{hyperref}

\SetCommentSty{mycommfont}
\usepackage{enumitem}
\usepackage{lipsum}
\usepackage{stfloats}
\usepackage{ulem}
\usepackage[numbers]{natbib}
\usepackage[usenames,dvipsnames]{xcolor}
\usepackage{comment}
\usepackage{flushend}
\usepackage{multirow}

\usepackage{tikz}
\usetikzlibrary{fit, shapes.misc}

\newcommand{\circnum}[1]{\raisebox{0.2ex}{\textcircled{\raisebox{-0.2ex}{#1}}}}
\newcommand{\limebox}[1]{\setlength{\fboxsep}{0pt}{\colorbox{lime}{#1}}}
\newcommand{\pinkbox}[1]{\setlength{\fboxsep}{0pt}{\colorbox{yellow}{#1}}}
\usepackage{CJKutf8}
\definecolor{bleudefrance}{rgb}{0.19, 0.55, 0.91}
\definecolor{awesome}{rgb}{1.0, 0.13, 0.32}
\definecolor{darkgreen}{rgb}{0.0, 0.65, 0.0}
\definecolor{babyblue}{rgb}{0.29, 0.75, 0.93}

\definecolor{darkorange}{rgb}{1.0, 0.55, 0.0}

\begin{document}
\title{IKSel: Selecting Good Seed Joint Values for Fast Numerical Inverse Kinematics Iterations}
\author{Xinyi Yuan, Weiwei Wan$^{*}$, Kensuke Harada
\thanks{Department of System Innovation, Graduate School of Engineering Sci-
ence, Osaka University, Toyonaka, Osaka, Japan.}
\thanks{$^{*}$Contact: Weiwei Wan, {\tt\small wan@sys.es.osaka-u.ac.jp}}}
\markboth{Under Review by a Robotics Journal, 2025.}
{Yuan \MakeLowercase{\textit{et al.}}: Selecting Good Seed for Numerical Inverse Kinematics}
\maketitle

\bstctlcite{IEEEexample:BSTcontrol}
\begin{abstract}
This paper revisits the numerical inverse kinematics (IK) problem, leveraging modern computational resources and refining the seed selection process to develop a solver that is competitive with analytical-based methods. The proposed seed selection strategy consists of three key stages: (1) utilizing a K-Dimensional Tree (KDTree) to identify seed candidates based on workspace proximity, (2) sorting candidates by joint space adjustment and attempting numerical iterations with the one requiring minimal adjustment, and (3) re-selecting the most distant joint configurations for new attempts in case of failures. The joint space adjustment-based seed selection increases the likelihood of rapid convergence, while the re-attempt strategy effectively helps circumvent local minima and joint limit constraints. Comparison results with both traditional numerical solvers and learning-based methods demonstrate the strengths of the proposed approach in terms of success rate, time efficiency, and accuracy. Additionally, we conduct detailed ablation studies to analyze the effects of various parameters and solver settings, providing practical insights for customization and optimization. The proposed method consistently exhibits high success rates and computational efficiency. It is suitable for time-sensitive applications.
\end{abstract}

\def\abstractname{Note to Practitioners}
\begin{abstract}
The proposed IK solver serves as the default IK solver of our WRS Robot Planning and Control System (https://github.com/wanweiwei07/wrs). The core implementation of the method is located in the $\mathtt{ik\_sel.py}$ file within the $\mathtt{wrs.robot\_sim.\_kinematics}$ package. For rapid prototyping, we provide a manipulator interface class in $\mathtt{wrs.robot\_sim.manipulators.manipulator\_interface}$. Practitioners can define specialized manipulators by implementing the interface and then invoking the inherited $\mathtt{ik()}$ member function to solve the manipulator's IK using the proposed method. Users may adjust the maximum allowed number of re-selection attempts by modifying line 173 in the $\mathtt{ik\_sel.py}$ file to achieve an optimal balance between success rate and computation efficiency based on their requirements. For integration with other simulation systems, practitioners can refer directly to the structure and logic provided in the standalone implementation of the $\mathtt{ik\_sel.py}$ file. This file is fully aligned with the methods presented in this paper and offers a clear, modular reference for reproducing the proposed selection strategy.
\end{abstract}


\section{Introduction} 

\begin{figure}[tbp]
    \centering
    \includegraphics[width=\linewidth]{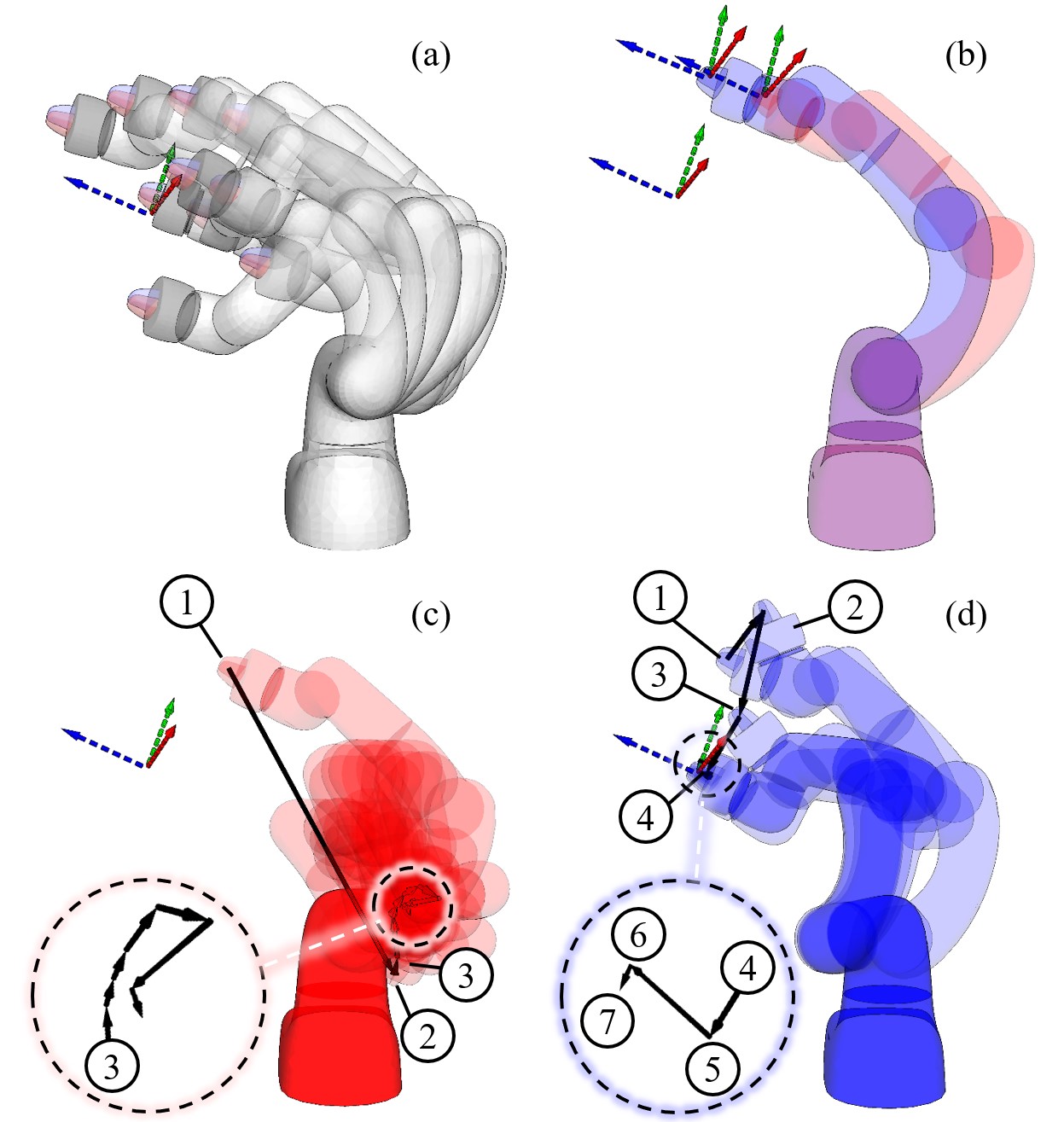}
    \caption{Comparison of success and failure seed joint configurations for numerical IK: (a) Several candidate seed configurations with small workspace deviations from the target Tool Center Point (abbreviated as TCP, indicated by the dashed coordinate frame). (b) Two representative seeds were selected for comparison. Based on the Euclidean distance metric in the workspace, the TCP of the red seed is closer to the target TCP than that of the blue seed. (c) However, without applying a scaling coefficient, the red seed significantly deviates after the first numerical iteration. (d) In contrast, the blue seed rapidly converges to the target TCP within only four iterations, without requiring any coefficient adjustment. This paper proposes a method to effectively identify and select seeds similar to the blue one, thereby reducing iteration counts and enhancing IK solving efficiency.}
    \label{workflow}
\end{figure}  

\IEEEPARstart{C}{urrently}, it is known that 6 Degree-of-Freedom (DoF) robotic arms always possess analytical IK solutions \cite{lee1988new}. However, does this imply numerical IK methods are no longer necessary? The answer remains negative. In practical implementations across various robotic arm manufacturers, numerical methods still play a central role in solving IK for many 6 DoFs arms \cite{chiaverini1994review}. Although analytically solvable in theory, deriving general closed-form solutions for these robots is highly challenging. Even symbolic computation tools frequently fail due to computational complexity. Consequently, numerical solutions remain crucial for practical applications. Moreover, robotic arms with 7 or more DoFs, which are increasingly prevalent to overcome mechanical constraints and enhance flexibility -- particularly in welding applications -- rely heavily on numerical IK methods. Numerical methods can seamlessly integrate initialization strategies and optimization objectives, thereby offering substantial advantages. It also allows handling null-space projections and multiple tasks for the high-DoF robots. Therefore, research into numerical IK methods remains both important and challenging. 

Motivated by these considerations, this paper revisits numerical IK with the aim of leveraging modern computational resources and refined computational strategies to significantly reduce computational costs. Our objective is to enable numerical methods to approach the computation speed and efficiency of analytical solutions. To achieve this goal, we propose an improved seed selection approach for numerical iterations. The core idea is intuitive: employing a KDTree structure to select better initial seeds, thus reducing iteration counts and achieving rapid convergence. Nonetheless, this intuitive concept is non-trivial in practice. A simple nearest-neighbor selection algorithm cannot accurately pinpoint seeds that lead to valid solutions, let alone ensure fewer iterations. Fig. \ref{workflow}(a) and (c) illustrate this non-trivial seed selection problem.

To solve the problem, we carefully investigate the relationship between the nearest-neighbor seeds in the KDTree and the inverse solutions and propose a method to rank and select optimal seeds based on the linearity of individual models. Specifically, we select the seed that minimizes the required joint value adjustments after multiplication by the present Jacobian inverse matrix, under the condition that the workspace difference between the seed TCP and the target TCP is small. If a selected seed fails, the next iteration re-selects a seed by prioritizing the one whose joint configuration is farthest from the previously failed seed. By adopting the proposed selection and re-selection method, it becomes unnecessary to apply scaling coefficients to joint adjustments, thereby significantly reducing the required number of iterations and improving computational efficiency. Fig. \ref{workflow}(d) exemplifies the numerical iteration process using a seed configuration selected by the proposed method.

It is worth noting that while our method relies on KDTrees and is data-driven, it does not utilize neural networks. This aligns with our goal of fast computation. While many learning-based IK solvers have shown strong performance, their accuracy and speed are heavily dependent on network design and training quality. The learning-based methods are often impractical for time-sensitive applications. From this perspective, our approach maintains relevance even with the current AI-driven research landscape. It provides a computationally efficient alternative data-driven solution without the complexity and inference costs associated with neural networks.

In the experiments, we compared our proposed method with both traditional and learning-based IK solvers in terms of success rate, time efficiency, and accuracy. Results demonstrate that the proposed seed selection and re-selection methods significantly improve the success rate of the numerical solver without sacrificing the computational speed. Besides, comprehensive ablation studies are conducted to provide insights into the influence of parameter tuning and solver settings, facilitating practical customization according to specific application requirements and robotic systems.

\section{Related Work} 
The primary IK solving methods are broadly categorized into analytical, numerical, and learning-based approaches.

\subsection{Analytical Methods}
Analytical methods aim to find explicit closed-form solutions based on the geometry and constraints of the mechanism. Early studies began with special configurations and gradually extended to general 6R manipulators. For example, Pieper et al. \cite{pieper1969kinematics} obtained closed-form IK solutions for 6 DoFs manipulators with three revolute joints intersecting at a point. The remaining joints could either be rotary or prismatic. Tsai et al. \cite{tsai1985solving} hypothesized that the IK problem for general 6R manipulators has at most 16 solutions based on empirical observations. Lee et al. \cite{lee1988new} validated the hypothesis using a new vector theory and derived a 16th-order polynomial in the tangent of the half-angle. Raghavan et al. \cite{raghavan1990kinematic}\cite{raghavan1993inverse} and Lee et al. \cite{lee1991complete} further presented simpler solutions by applying dialytic elimination. Kohli et al. \cite{kohli1993inverse} and Manocha et al. \cite{manocha1994efficient} reduced the problem to finding the eigenvalues of a 16$\times$16 or 12$\times$12 matrix, respectively, where each eigenvalue corresponds to a root in the 16th-order polynomial. Rudny et al. \cite{rudny2014solving} employed Bernstein elimination to find the intersections of bivariate polynomials, thereby avoiding tangent half-angle substitution. More recent methods like IKFast \cite{diankov2010automated} produced closed-form solutions for kinematic chains employing hinge and prismatic one-degree-freedom joints. Similarly, IKBT \cite{zhang2017ikbt} constructed Behavior Trees (BT) to symbolically encode and solve IK equations within minutes. They significantly reduced the derivation time. 

Although analytical methods offer precise and computationally efficient solutions, they can be challenging or infeasible to derive even for some 6R manipulators and may face additional difficulties with redundant or high-DoF systems \cite{aristidou2018inverse}. Some semi-analytical approaches integrate numerical techniques into proposed methods, thus broadening the applicability while maintaining accuracy. IK-Geo \cite{elias2025ik}, for example, provides precise closed-form or numerical-based solutions for 6 DoFs robots by decomposing the IK into several sub-problems.

The persistent challenges of analytical methods make it significant to study numerical methods. In particular, general numerical IK methods that do not require preprocessing for each specific configuration represent a promising research topic. However, the primary bottleneck of numerical methods is low computational efficiency and success rates, which serves as the main motivation for this study. We will review existing numerical methods in the following subsection.

\subsection{Numerical Methods}

Numerical methods iteratively solve IK by minimizing the distance between the manipulator's current TCP pose and a target TCP pose. These methods exhibit flexibility in accommodating various manipulator kinematic configurations.

The core of numerical IK methods is the Jacobian matrix. Under small joint angle changes, kinematic mappings can be treated linearly, and numerical IK methods iteratively adjust joint angles using the inverse of the Jacobian matrix to reduce pose errors until convergence \cite{whitney1969resolved}. Previous studies have extensively investigated efficient computation or updating strategies for the Jacobian matrix \cite{orin1984efficient}. Numerous approaches have also been developed to circumvent singularities during matrix inversion \cite{nakamura1986inverse}\cite{wampler1986manipulator}\cite{sugihara2011solvability}, or to efficiently compute generalized inverses \cite{mayorga1992fast}. Additionally, some researchers have proposed using an approximate Hessian matrix instead of the Jacobian matrix to accelerate numerical convergence \cite{lenarvcivc1985efficient}\cite{zhao1994inverse}. Studies have also explored methods to handle joint limitations effectively during numerical iterations \cite{sekiguchi2021numerical}\cite{huang2016clamping}.

Despite the adaptability and the many previously developed algorithms for acceleration and stability, numerical IK methods remain sensitive to initial seed selections. Different initial seeds can lead to different adjusting directions and significantly influence convergence speed. Consequently, applying seed-selection techniques can be especially important for numerical IK methods. Previously, \citet{farzan2013dh} accelerated numerical IK convergence by generating Jacobian matrices in parallel and subsequently selecting the optimal joint configuration in each iteration. This work also focuses on selecting optimal seeds for numerical methods. We first use KDTree to identify seed candidates based on workspace
proximity. Then, we sort the seed candidates by looking one step forward to their joint space adjustment and prioritize the one requiring minimal adjustment for numerical iteration. If the chosen seed fails to yield a valid solution, we implement a re-selection strategy, prioritizing candidates with the largest joint configuration differences from the previously failed seeds. The seed selection procedure based on joint space adjustment enhances the likelihood of rapid convergence, while the re-selection strategy helps effectively circumvent local minima and joint limits. 

\subsection{Learning-Based Methods}

Neural networks exhibit remarkable capabilities in approximating complex, non-linear relationships within high-dimensional data spaces. However, the inherently one-to-many mapping characteristic of IK problems complicates convergence for neural IK methods \cite{demby2019study}, as multiple joint configurations can correspond to a single target pose. To address this issue, previous research typically formulated IK as a single-mapping problem between inputs and desired outputs \cite{csiszar2017solving}\cite{xia2001dual}\cite{oyama2001inverse}. For instance, Lendaris et al. \cite{lendaris1999linear} proposed an augmented linear Hopfield network incorporating a feedforward layer to approximate the Moore–Penrose generalized inverse of the Jacobian. D'Souza et al. \cite{d2001learning} embedded null-space optimization into the network's cost function to resolve redundancy. Oyama et al. \cite{oyama2001inverse} developed a modular neural network system comprising multiple expert networks to approximate continuous IK function regions and selected the most suitable expert networks for kinematic control.

Beyond the above seminal work, recent studies have also explored using Deep Reinforcement Learning (DRL) solutions for IK. They particularly focused on addressing key challenges such as computational complexity and redundancy. For example, \citet{phaniteja2017deep} utilized Deep Deterministic Policy Gradient (DDPG) to solve dynamically stable IK for a 27 DoFs humanoid robot, achieving 90\% end-effector accuracy. Similarly, \citet{guo2019reinforcement} applied DDPG to solve IK for 5 DoFs manipulators. \citet{malik2022deep} proposed a Deep Q-Network (DQN)-based IK solver for 7 DoFs manipulators, where the network takes specified Cartesian end-effector trajectories as input and outputs discrete joint space actions. Compared to DDPG, which has a continuous action space, the simpler architecture of DQN enables easier and faster implementation. For the Proximal Policy Optimization (PPO)-based method, \citet{perrusquia2021multi} utilized the Multi-Agent Reinforcement Learning (MARL) framework that treats each joint in the redundant robot as an independent agent. \citet{zhang2022simulation} further improved the PPO algorithm with hierarchical rewards and integrated visual inputs from the simulation environment. More recently, \citet{zhao2024inverse} developed a Multi-agent PPO (MAPPO) IK solver with real-time obstacle-aware reward mechanisms to achieve singularity-robust IK solving for 6 DoFs manipulators in changeable environments.

Generative models have emerged as another promising tool to learn the intrinsic ambiguity of IK problems. Unlike deterministic approaches, generative models output samples from a probability distribution, making them naturally suited to represent the multiple feasible joint configurations that correspond to a single end-effector target. Previously, Ardizzone et al. \cite{ardizzone2018analyzing} and Kruse et al. \cite{kruse2021benchmarking} utilized the Invertible Neural Network (INN) to achieve Normalizing Flows on a simple 2D space IK problem, where the model outputs the posterior distribution of joint angles conditioned on the end-effector goal pose. Based on previous works, IKFlow \cite{ames2022ikflow} extended the approach to the 7 or more DoFs arm workspace while improving prediction precision and training stability. Additionally, Ren et al. \cite{ren2020learning} applied modified Generative Adversarial Networks (GANs) to learn IK and Inverse Dynamics (ID) using limited real-world data. Ho et al. \cite{ho2023deep} introduced the posture index learned by the Variational Auto-Encoder (VAE) to probabilistically map poses to distinct joint solutions and obtained IK solutions for the 7 DoFs Franka Emika Panda robot. Despite their effectiveness, generative IK solvers require robot-specific training. Recent studies like Limoyo et al. \cite{limoyo2024generative} attempted to address the limitation by incorporating distance-geometric robot representation into the graph-based generative model and successfully solved IK problems for unseen robots.
 
Although learning-based methods are a popular research focus, they face challenges such as accuracy and generalization when applied to IK problems. Computational speed is also a concern, as repeated convolutions within neural networks are often inefficient on CPUs, while GPUs' advantages only emerge for networks with a sufficient number of layers and model complexity. Additionally, the data transfer overhead between CPUs and GPUs can further reduce overall efficiency. Generative models typically require computation time exceeding tens of milliseconds, limiting their applicability in time-sensitive scenarios. For these reasons, this study employs KDTree-based methods for efficient data handling. The experimental section provides a detailed comparative analysis to demonstrate the strengths and limitations of the proposed method in terms of success rate, time efficiency, and accuracy. Nonetheless, we also show that under conditions with less stringent time constraints, employing generative models for IK problems holds significant potential.

\section{Selecting Seeds using KDTree}

The convergence and speed of numerical IK are highly sensitive to initial seed selection. In this section, we propose a framework that utilizes the KDTree structure to store seed joint configurations and select seeds considering the linearity in their nearby IK function space. In detail, the selection process involves two steps: (1) utilizing a KDTree to identify seed candidates based on workspace proximity and (2) sorting candidates by joint space adjustment and attempting numerical iterations with the one requiring minimal adjustment.

The construction of a KDTree begins with uniform sampling of the manipulator's joint configuration space to ensure representative coverage. For each sampled joint value $\mathbf{q_{\text{i}}}$, the corresponding TCP position and rotation $\mathbf{(p_{\text{i}},w_{\text{i}})}$ are computed using forward kinematics following
\begin{equation}
\mathbf{x_{\text{i}}} = 
    \begin{bmatrix}
\mathbf{p_{\text{i}}} \\
\mathbf{w_{\text{i}}}
\end{bmatrix} = f(\mathbf{q_{\text{i}}}),
\end{equation}
Here, $\mathbf{x_{\text{i}}}$ is the TCP pose, $\mathbf{p_{\text{i}}}$ is the 3-dimension position, $\mathbf{w_{\text{i}}}$ is the rotation vector, $\mathbf{q_{\text{i}}}$ is the joint configuration and function $f()$ stands for the forward kinematics computation\footnote{Notably, the rotation vector ($\mathbf{w}$) is one of several representations for the end-effector orientation. We also evaluated alternatives such as the flattened rotation matrix,  Euler angles, and quaternions. However, ablation experiments showed no significant differences among these representations. Dual quaternion \cite{beeson2015trac} is recommended for more precise modeling if concerned.}. The sampled $\mathbf{q_{\text{i}}}$ and the corresponding computed $\mathbf{p_{\text{i}}}$ and $\mathbf{w_{\text{i}}}$ are then used to construct the KDTree. As a result, each node in the KDTree stores a TCP pose ($\mathbf{x_{\text{i}}}$) and its correspondent joint configuration ($\mathbf{q_{\text{i}}}$).

\subsection{Intuitive Seed Selection Method and Challenges}

Given a target TCP pose $\mathbf{x_{\text{t}}}$ in the workspace, we can query the KDTree to find the nearest neighbor of $\mathbf{x_{\text{t}}}$ using the $L_\text{2}$ norm shown below.
\begin{equation}
\mathbf{q_\text{seed}} = \arg\min_{\mathbf{q} \in \mathcal{Q}} ||\mathbf{x_{\text{t}}} - \mathbf{x_{\text{i}}}||^2,
\label{eq:nearest_neighbor}
\end{equation}
Here, $\mathcal{Q}$ denotes the joint space in the KDTree. As a seed selection strategy for numerical IK, a simple and direct approach is to choose the joint configuration whose TCP pose yields the smallest $L_\text{2}$ norm difference from the target pose as $\mathbf{q_{\text{seed}}}$. This seed can then be provided to a Newton-Raphson solver for computing an accurate IK solution. If the iteration converges, the algorithm terminates and returns the result. Otherwise, it reports failure.

Alternatively, we do not necessarily need to restrict the selection to a single seed. We can select a subset of candidate seeds whose workspace distances are within a predefined threshold $\delta$:
\begin{equation}
\mathcal{Q}_\delta = \{ \mathbf{q} \in \mathcal{Q} \mid ||\mathbf{x_{\text{t}}} - \mathbf{x_{\text{i}}}||^2 \leq \delta \}.
\label{eq:thresholded_candidates}
\end{equation}
Within this subset, the candidate seeds can be sorted based on a given metric (e.g., $L_\text{2}$ norm or local linearity properties discussed in the next section). The algorithm then traverses this ordered list, using each seed $\mathbf{q_{\text{seed}}}$ as an initialization for numerical iteration. The first seed that successfully leads to a valid IK solution is returned as the final result. If all candidates in $\mathcal{Q}_\delta$ fail to converge to a valid solution, the algorithm declares failure.

The above intuitive seed selection method, which involves pre-storing a KDTree using an $L_\text{2}$ norm metric and selecting optimal seeds for iterative computation, is simple and direct. However, the success rate of this method is relatively low. The $L_\text{2}$ metric does not guarantee that the selected seed will effectively and rapidly converge to an accurate numerical solution. This limitation arises because the metric is defined in the workspace, whereas variations in the joint space can differ substantially. Therefore, a more delicate seed selection strategy is necessary to improve convergence efficiency and robustness.

\subsection{Seed Selection Considering Linearity}

Instead of metrics in the workspace, we explore a joint space criteria to better capture the underlying shape of the kinematic function. Specifically, we develop a ranking method to select the optimal seed among KDTree candidates based on joint space linearity. This approach is motivated by the following insight.

During numerical iteration, joint value adjustments should be kept as small as possible, as they theoretically represent velocity rather than discrete differences. Smaller adjustments lead to more accurate approximations when using the Jacobian matrix computed at the current joint configuration. A common practice is to apply a scaling coefficient to limit the magnitude of these adjustments. In the absence of such a coefficient, the updates may oscillate and lead to instability. However, the use of a scaling coefficient slows down convergence. It limits the amount of adjustment per iteration, thereby slowing down convergence and reducing the efficiency of the numerical solver. Since our method is supported by KDTree, when the TCP difference is within a certain threshold, we can select a seed that inherently requires smaller joint space adjustments from the data saved in the tree, thereby eliminating the need for additional scaling.

Some researchers may be concerned that the idea could lead to a strong dependency on the density of the KDTree, as a theoretical justification for omitting a scaling factor only holds when the KDTree is sufficiently dense and the workspace distance remains negligibly small. However, in practice, this is not necessarily the case. Seeds that facilitate efficient numerical iteration are often located in regions that locally satisfy Lipschitz continuity or, more strictly, exhibit local linearity. In such regions, even if the initial workspace deviation is relatively large, the iteration is still likely to converge quickly to the solution. Conversely, seeds that are difficult to iterate to a solution tend to appear in regions with high differential variation, where numerical iteration is more prone to instability. When the workspace difference and its corresponding joint space adjustments are equally small, it suggests a locally well-behaved linear region, increasing the likelihood of rapid convergence. Note that this is not a strict guarantee. It only implies a higher probability of successful iteration without scaling. The most effective seeds are highly likely to be found among candidate solutions where the joint space adjustments are relatively small.

The above insight motivates us to develop the following seed selection method. In addition to the joint values and TCP pose, we pre-compute the pseudo-inverse of the Jacobian matrices corresponding to the joint values via Singular Value Decomposition (SVD) and store the result within the KDTree structure. Specifically, given a Jacobian matrix with its SVD decomposition $\mathbf{J = U\Sigma V^\text{T}}$, the pseudo-inverse is defined as $\mathbf{J^+ = V\Sigma^+ U^\text{T}}$, where each element in $\mathbf{\Sigma^+}$ is the reciprocal of the non-zero singular values in $\mathbf{\Sigma}$\footnote{When the singular values approach zero, their reciprocals become extremely large, which leads to significant numerical instability. To solve the problem, we can set a relative regularization threshold to cut off small singular values. The threshold can be determined by the product of the largest singular value and a specified constant parameter. The regularization process effectively improves the numerical stability and the robustness of the calculation.}. For a seed $\mathbf{q_{\text{seed}}}$, the approximated linear update from the seed joint value is calculated as:
\begin{equation}
\Delta \mathbf{q_{\text{seed}}} = \mathbf{J}_{\text{seed}}^+ \cdot 
\Delta 
\begin{bmatrix}
\mathbf{p} \\
\mathbf{w}
\end{bmatrix}
\label{eq:delta_q}
\end{equation}
where $\mathbf{J}_{\text{seed}}^+$ is the Pseudo-Inverse (PINV) of the Jacobian matrix at the seed. During the iteration process, we select seeds that are not only closer to the target pose in workspace but also have a smaller $\Delta \mathbf{q}$ in joint space to align with the insight discussed above, thereby improving both the efficiency and success rate of the Newton-Raphson solver.

As a result, each node in the KDTree stores the joint configuration ($\mathbf{q}$), the end-effector target state ($\mathbf{x}$), and the pseudo-inverse of the Jacobian matrix ($\mathbf{J}^{+}$). To improve the robustness and efficiency of seed selection, we adopt a two-stage filtering strategy. First, we identify a subset $\mathcal{Q}_\delta$  of candidate seeds whose workspace distances from the target pose fall within a predefined threshold following the equation \eqref{eq:thresholded_candidates}. This constraint ensures that only seeds with sufficiently small workspace deviations are considered, reducing the likelihood of selecting initial configurations that are too far from the desired target. Second, we prioritize seeds within $\mathcal{Q}_\delta$ based on the magnitudes of their joint space adjustments. We define the joint space variation $L_\text{2}$ norm as the evaluation metric
\begin{equation}
||\Delta \mathbf{q_\text{seed}}||_2 = \sqrt{\sum_{\text{i}=1}^{dof} (\Delta \mathbf{q_\text{i}})^2},
\label{eq:l2_norm}
\end{equation}
and rank the candidates in $\mathcal{Q}_\delta$ based on this metric. In this way, we select the seed that requires minimal joint space adjustments for numerical iteration. This two-stage selection strategy improves convergence efficiency and reduces failure rates by increasing the likelihood that the initial seed lies within a linear region of the kinematic function.

\section{Numerical Iteration and Re-Selection}

\begin{figure}[!htbp]
    \centering
    \includegraphics[width=\linewidth]{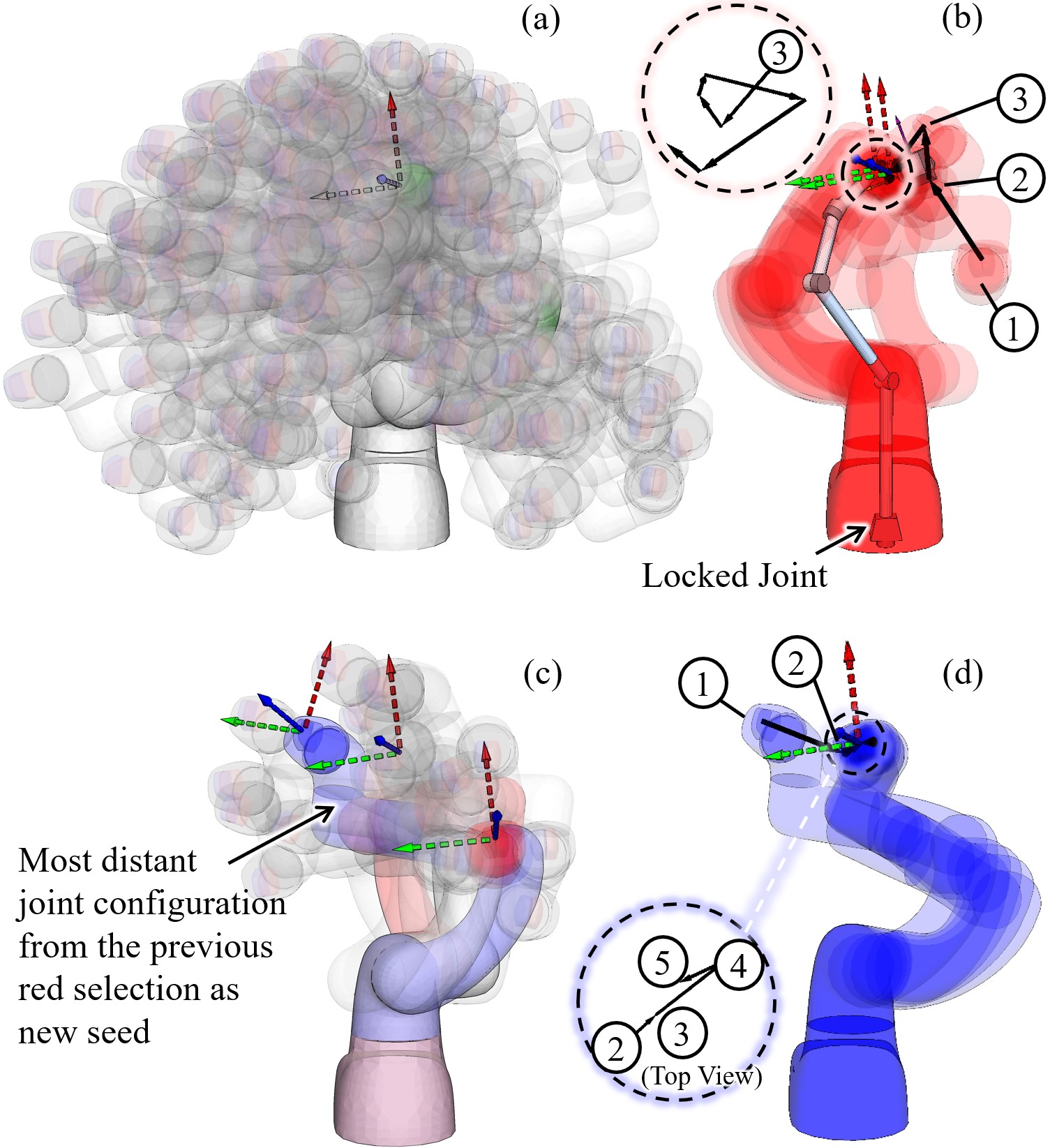}
    \caption{(a) 200 candidate seed configurations. The green robot with a dashed coordinate frame is the solution. (b) The seed labeled with \circnum{1} has the smallest joint space adjustment and is selected for numerical iteration. However, after several steps, the robot's first joint approximates its joint limit and fails to converge. (c) The most distant joint configuration from the failed seed is re-selected for a new numerical solution. The re-selection is performed within the top 20 candidates with the smallest joint space adjustment in order to impose a strong linearity condition near the new selection and thus increase the likelihood of fast convergence. (d) The re-selected seed finally converges to the target.}
    \label{joint_lock}
\end{figure}

After filtering and sorting the seed candidates by their combined workspace proximity and joint space linearity, the success rate of numerical IK can be effectively improved. It is important to acknowledge, however, that this approach is inherently heuristic and lacks a mathematically rigorous guarantee. Failure cases may still arise, particularly due to joint motion range limitations. The solver may become stuck when specific joints approach their motion limits after several iterations. For instance, the \circnum{1} seed shown in Fig. \ref{joint_lock}(b) has the minimum joint space adjustment among the candidates shown in Fig. \ref{joint_lock}(a). However, the robot's first joint approaches its limit after several numerical iterations and fails to converge. Another significant cause of failure is the objective to minimize the number of iterations and reduce computational time costs. Sometimes, the selected seed may have a low convergence rate. Although it falls in a linear region, the seed may still fail to meet the precision requirements within a predefined iteration threshold (e.g., 10 iterations).

In conventional literature, the primary approaches for addressing joint limit violations during numerical iteration are the Weighted Jacobian methods \cite{ito2017kinematic} and Null Space methods \cite{nakamura1987task}. Both methods fundamentally rely on defining a metric function to quantify proximity to joint limits and subsequently adjusting joint updates based on this metric. However, these methods typically entail extended iteration processes to guide the manipulator to circumvent joint limits and converge to the target configuration. This iterative overhead conflicts with the objective of minimizing iteration times and enhancing numerical efficiency. For this reason, these conventional approaches are not well-suited for our framework. Alternative strategies are required to achieve efficient convergence while addressing joint limit constraints. 

To deal with numerical iteration failures, we propose a re-selection strategy that resets the initial value to another candidate most likely to succeed based on previous failures. Previous studies, such as \citet{sekiguchi2021numerical}, have suggested that modifying the end-effector's rotation can help escape local minima (deadlocks) during numerical IK iterations. Building on this insight, when the numerical iterations encounter failures, our algorithm re-selects the most distant seed from the previously failed ones for a new numerical iteration. This approach is based on the premise that a seed farther from the failed configuration is less likely to be constrained by joint limits and, therefore, has a greater potential to avoid deadlock regions.

\begin{figure}[!htbp]
    \centering
    \includegraphics[width=\linewidth]{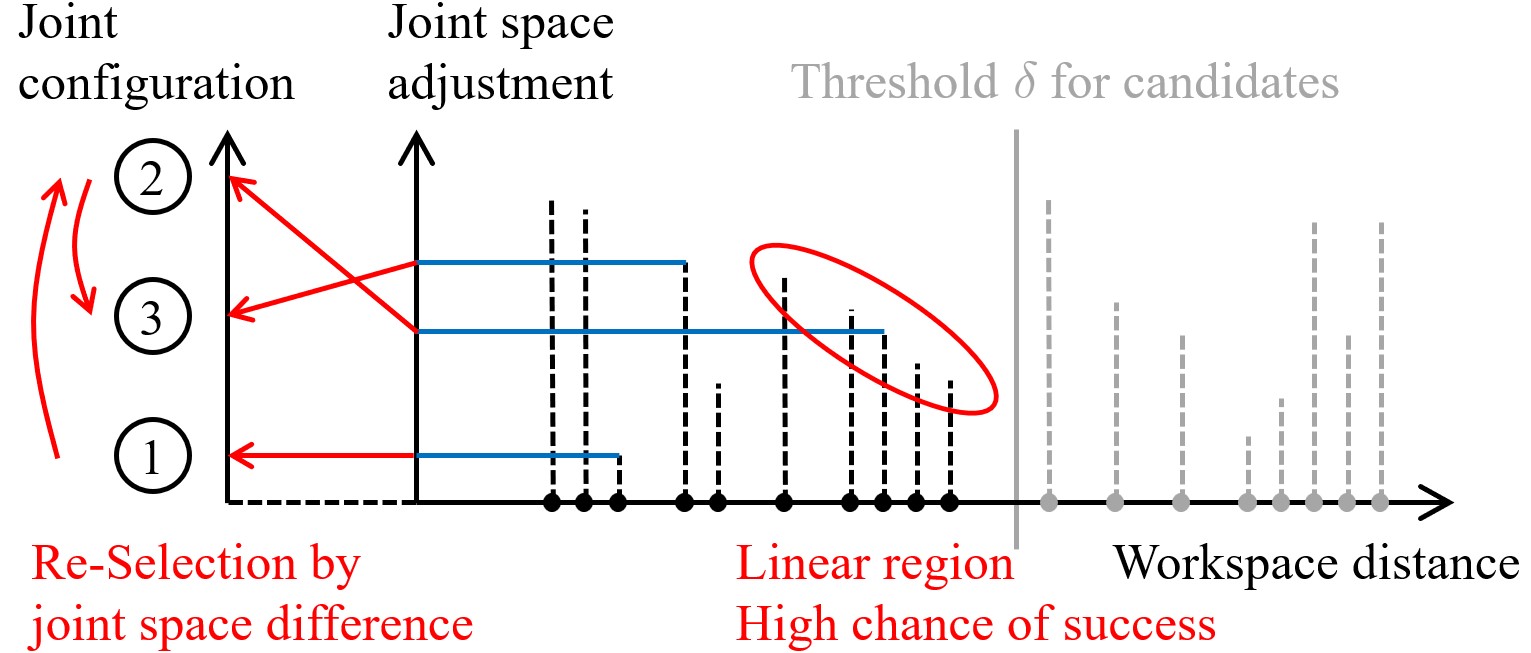}
    \caption{Conceptual sketch of the re-selection strategy.}
    \label{fig_binary}
\end{figure}

From another perspective, our re-selection strategy can be regarded as a form of binary search. Fig. \ref{fig_binary} illustrates the concept. Within the set of candidate seeds whose workspace distances fall below the predefined threshold $\delta$, our strategy first selects the candidate with the smallest joint adjustment as the initial seed for numerical iteration. If this seed fails to converge, the strategy re-selects the candidate that has the most different joint configuration from the previously failed seeds and restarts the numerical iteration. In the illustrated iteration and re-selection process shown in Fig. \ref{fig_binary}, the solver initially selected candidate \circnum{1}, which exhibited the smallest joint space adjustment. Although this candidate had minimal joint adjustment, it failed to converge to a solution that met the required precision. Subsequently, the solver re-selected candidate \circnum{2}, which had the largest difference in joint configuration from the previous candidate, and initiated a new iteration. However, this attempt also failed to yield a valid solution. Finally, the solver selected candidate \circnum{3}, which was significantly different in joint configuration from both previous candidates. This re-selection led to successful convergence.

Note that an important implementation detail in the re-selection process is to restrict the comparison range to seed candidates whose joint space adjustments fall below a specified threshold. This would continuously enforce a strong linearity condition near the new seed, which in turn tends to yield a higher success rate.

In summary, the efficiency of our KDTree-based method is primarily attributed to three key aspects. First, a workspace distance threshold is applied to determine the subset of candidate seeds. This effectively narrows down the search space by filtering out candidates that are too distant from the target in the workspace. Second, within this subset, candidates are ranked based on joint space adjustment, and the seed with the smallest adjustment is selected for the initial numerical iteration. This prioritization enhances the likelihood of rapid convergence by selecting seeds that are in the linear region of the joint space manifold. Third, when a selected seed fails to converge, a re-selection strategy is employed, where the next seed is selected as the one that has a maximal difference in joint configuration from the previously failed seeds. This helps avoid local minima and joint limits, improving the robustness of the solver. The computational cost of solving an IK problem using the proposed method can be expressed as the sum of the KDTree search cost, the sorting cost, and the product of the number of re-selection attempts and the iteration count per attempt. The seed selection strategy allows for setting relatively small limits for both the iteration count and the number of re-selection attempts while maintaining a high success rate. The small limits help significantly reduce overall computational costs. Detailed comparisons and analysis of parameter choices will be presented in the experimental section.

\section{Experiments and Analysis} 

\begin{figure}[tbp]
    \centering
    \includegraphics[width=\linewidth]{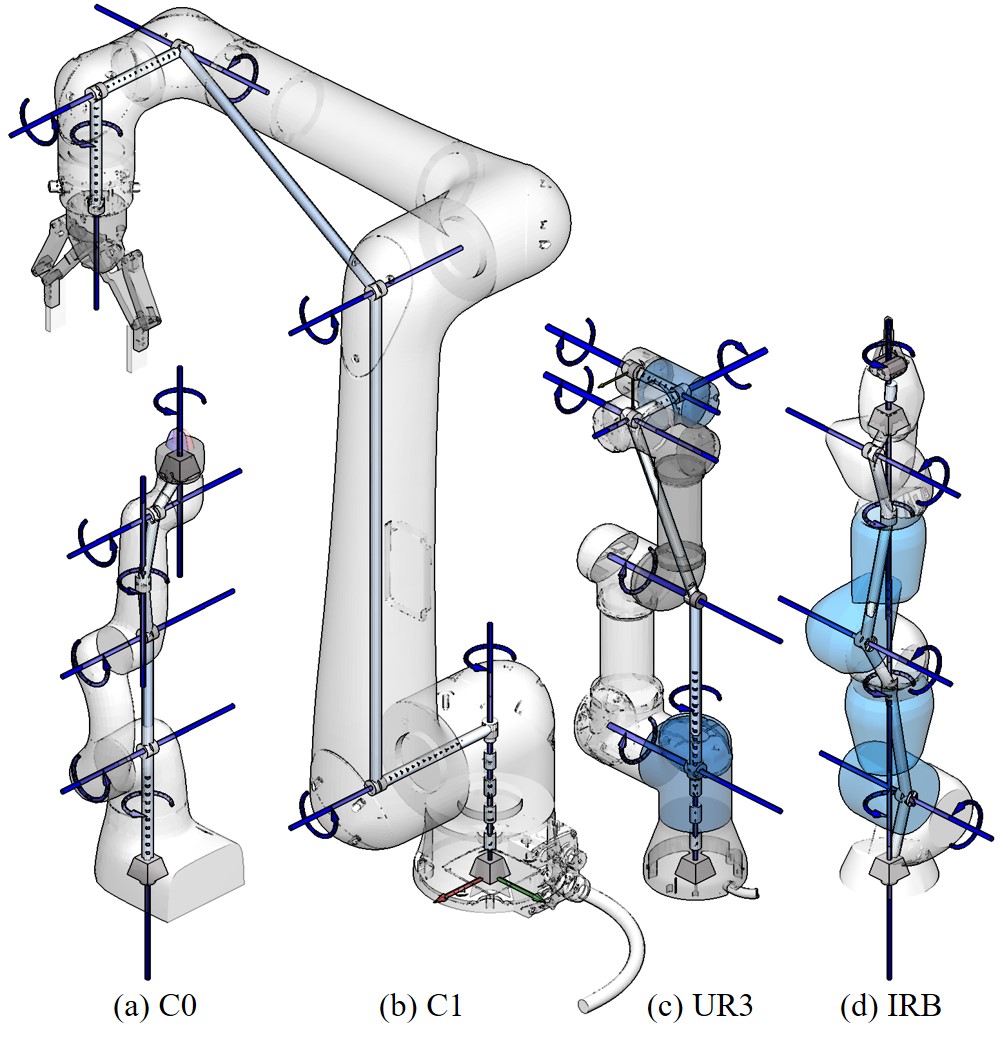}
    \caption{Kinematic structures of the manipulators used in the experiments. The dark blue line segments represent the motion axes of the joints. The dark blue circular arrows represent the rotational directions.}
    \label{robo_kine}
\end{figure}

In the experimental section, we first conduct an ablation study to analyze the efficiency and success rate of the proposed method under different parameter settings. Subsequently, we perform a benchmark comparison between the proposed method and several conventional approaches, as well as learning-based methods, to highlight the advantages and limitations of the proposed approach. All experiments were conducted using the WRS robotic system on a PC with a 13th Gen Intel(R) Core(TM) i7-13700KF CPU, an NVIDIA GeForce RTX-4090 GPU, and 64 GB of RAM, running an Ubuntu 22.04 operating system. The implementing language is Python 3.10. All code can be found in our WRS simulation system, as mentioned in the ``Note to Practitioners'' sub-abstract. For the experimental evaluation, we selected the following four robotic manipulators. They are illustrated in Fig. \ref{robo_kine} for reference.
\begin{itemize}[leftmargin=1em]
 \renewcommand{\labelitemi}{$\triangleright$}
    \item \uline{CVR038 by Denso (C0):} The Cobotta Mini collaborative robot with 6 DoFs. Its kinematic structure is characterized by parallel alignment of axes 2 and 3, while axes 1 and 2, 3 and 4, and 5 and 6 are orthogonal. Explicitly expressing the closed-form IK equations of this structure is difficult.
    \item \uline{CVRB1213 by Denso (C1):} The extended-reach version of the Cobotta Pro collaborative robot. It has the same kinematic structure as the CVR038. The purpose of including it is to investigate the relationship between the sizes of the manipulator's workspace and the KDTree.
    \item \uline{IRB14050 by ABB (IRB):} One arm of the YuMi robot. It has 7 DoFs. 
    \item \uline{UR3 by Universal Robots (UR3):} A 6 DoFs collaborative robot features parallel alignment of axes 2, 3, and 4, with axes 1 and 2, 4 and 5, and 5 and 6 arranged orthogonally. This structure allows for a straightforward analytical representation.
\end{itemize}

\subsection{Ablation Study}

We first conduct an ablation study on the proposed method. Unless otherwise specified, the default setting involves sorting candidates based on joint space adjustment and performing one re-selection attempt upon failure. The default numerical solver uses the Damped Least Squares (DLS) method. Default candidate sorting is performed based on the nearest neighbors of the first 200 nodes in the KDTree. The default maximum number of allowed numerical iterations per attempt is 7.

\subsubsection{KDTree Sizes}
Table \ref{tab:ablation1} provides the variation of success rates and time costs with respect to different KDTree scales. For detailed information regarding the sample numbers and query file sizes corresponding to different KDTree scales, please refer to Table \ref{tab:kdtree-sizes} in the Appendix. The success rates steadily increase as the KDTree scales grow across all robots. It suggests that a more dense sampling of the joint space improves the effectiveness of nearest-neighbor seed queries. The improvements are more significant for the UR3 and C1 robots, which indicates that the method is sensitive to large joint space or complex IK solution space scenarios. In addition, Table \ref{tab:ablation1} also shows that changing the KDTree scales from small to medium increases the average inference time, as a larger database causes higher query and computation time. However, further increasing the KDTree scales to large leads to lower time costs. It implies that the success rate improvement surpasses the additional computational costs. Based on the ablation result the table, we can conclude that a larger KDTree storage database leads to better performance, especially under complex kinematics scenarios. 

\begin{table*}[t]
\centering
\caption{Ablation Study I: Influence of KDTree size}
\label{tab:ablation1}
\begin{threeparttable}
\begin{tabular}{lc@{\hskip 6pt}c@{\hskip 6pt}c@{\hskip 6pt}cc@{\hskip 9pt}c@{\hskip 9pt}c@{\hskip 9pt}c}
\toprule
\multirow{2.5}{*}{Scale} & \multicolumn{4}{c}{Success Rate (\%)} 
& \multicolumn{4}{c}{Time Cost (Mean / Std / Min / Max Time (ms))} \\ 
\cmidrule(lr){2-5} \cmidrule(lr){6-9}
& {C0} & {C1} & {UR3} & {IRB}
& {C0} & {C1} & {UR3} & {IRB} \\
\midrule
Small
& \pinkbox{81.61} & \pinkbox{83.49} & \pinkbox{35.30} & \pinkbox{93.99} 
& \pinkbox{1.74} / 0.75 / 0.86 / 4.99 & \pinkbox{2.22} / 0.99 / 0.88 / 7.99 & \pinkbox{3.39} / 1.04 / 0.87 / 6.65 & \pinkbox{1.68} / 0.69 / 0.97 / 5.61 \\
Medium
& 87.59 & 86.75 & 73.17 & 95.67
& 1.56 / 0.68 / 0.86 / 5.13 & 2.01 / 0.86 / 0.88 / 6.68 & 1.99 / 0.81 / 0.84 / 5.13 & 1.56 / 0.58 / 0.98 / 5.40 \\
Large
& \limebox{88.43} & \limebox{91.09} & \limebox{87.19} & \limebox{96.42}
& \limebox{1.54} / 0.68 / 0.89 / 6.38 & \limebox{1.89} / 0.73 / 0.93 / 6.91 & \limebox{1.79} / 0.87 / 0.90 / 7.09 & \limebox{1.55} / 0.53 / 1.03 / 6.92 \\
\bottomrule
\end{tabular}
\begin{tablenotes}
    \item ${}^{\text{Note 1~}}$The sample number and the corresponding KDTree query file sizes of Small, Medium, and Large scales are shown in Table \ref{tab:kdtree-sizes}.
    \item ${}^{\text{Note 2~}}$C0 -- CVR0308; C1 -- CVRB1213; IRB -- IRB14050.
    \item ${}^{\text{Note 3~}}$The text highlighted with a green background represents the best values, while the text highlighted with a yellow background indicates the worst values. All tables in this paper follow this coloring convention.
\end{tablenotes}
\end{threeparttable}
\end{table*}

\begin{table*}[t]
\centering
\caption{Ablation Study II: Variation in Numerical Solvers}
\label{tab:ablation2}
\begin{threeparttable}
\begin{tabular}{lc@{\hskip 6pt}c@{\hskip 6pt}c@{\hskip 6pt}cc@{\hskip 9pt}c@{\hskip 9pt}c@{\hskip 9pt}c}
\toprule 
 & \multicolumn{4}{c}{Success Rate (\%)} 
 & \multicolumn{4}{c}{Time Cost (Mean / Std / Min / Max Time (ms))} \\ 
\cmidrule(lr){2-5} \cmidrule(lr){6-9}  
 & C0 & C1 & UR3 & IRB 
 & C0 & C1 & UR3 & IRB \\
\midrule
DLS 
& 87.59 & 86.75 & 73.17 & 95.67
& \limebox{1.56} / 0.68 / 0.86 / 5.13 & \limebox{2.01} / 0.86 / 0.88 / 6.68 & \limebox{1.99} / 0.81 / 0.84 / 5.13 & \limebox{1.56} / 0.58 / 0.98 / 5.40 \\

PINV 
& 87.40 & 86.43 & 73.03 & 95.39 
& 1.60 / 0.71 / 0.86 / 6.12 & 2.09 / 0.89 / 0.91 / 8.15 & 2.05 / 0.85 / 0.87 / 7.16 & 1.60 / 0.60 / 0.78 / 6.39 \\
\(\text{PINV}_{\text{RR}}\) 
& \pinkbox{79.76} & \pinkbox{76.69} & \pinkbox{71.94} & \pinkbox{94.67} 
& \pinkbox{2.08} / 1.24 / 0.89 / 7.29 & \pinkbox{2.79} / 1.24 / 1.08 / 7.87 & 2.79 / 1.31 / 0.91 / 7.07 & \pinkbox{1.83} / 0.96 / 1.05 / 8.36 \\
CWLN 
& \limebox{91.24} & \limebox{88.95} & \limebox{80.34} & \limebox{96.63} 
& 1.95 / 1.32 / 0.87 / 9.10 & 2.44 / 1.34 / 0.92 / 9.05 & \pinkbox{2.83} / 1.54 / 0.90 / 8.93 & 1.70 / 0.97 / 0.99 / 5.87 \\
CWPI 
& 87.97 & 85.75 & 73.65 & 95.57 
& 1.78 / 0.91 / 0.92 / 7.37 & 2.27 / 0.97 / 0.98 / 8.79 & 2.25 / 0.96 / 0.95 / 7.27 & 1.71 / 0.60 / 1.08 / 7.04 \\
\bottomrule
\end{tabular}
\begin{tablenotes}
    \item ${}^{\text{Note~}}$The text highlighted with a green background represents the best values, while the text highlighted with a yellow background indicates the worst values. All tables in this paper follow this coloring convention.
\end{tablenotes}
\end{threeparttable}
\end{table*}

\begin{table*}[t]
\centering
\caption{Ablation Study III: Selection Metrics and Re-Selection Attempts}
\label{tab:ablation3}
\begin{threeparttable}
\begin{tabular}{l@{\hskip 11pt}c@{\hskip 5.5pt}c@{\hskip 5.5pt}c@{\hskip 5.5pt}c@{\hskip 11pt}c@{\hskip 8.5pt}c@{\hskip 8.5pt}c@{\hskip 8.5pt}c}
\toprule 
 & \multicolumn{4}{c}{Success Rate (\%)} 
 & \multicolumn{4}{c}{Time Cost (Mean / Std / Min / Max Time (ms))} \\ 
\cmidrule(lr){2-5} \cmidrule(lr){6-9}  
 & C0 & C1 & UR3 & IRB
 & C0 & C1 & UR3 & IRB \\
\midrule
Eq \eqref{eq:nearest_neighbor}
& 61.29 & 76.74 & 51.21 & 83.15 
& 2.00 / 0.87 / 0.84 / {\color{white}0}5.75 & \limebox{1.89} / 0.77 / 0.84 / {\color{white}0}5.53 & 2.05 / 1.06 / 0.83 / {\color{white}0}5.45 & 1.84 / 0.73 / 0.95 / {\color{white}0}5.38 \\

Eq \eqref{eq:l2_norm}
& \limebox{87.59} & \limebox{86.75} & \limebox{73.17} & \limebox{95.67}
& \limebox{1.56} / 0.68 / 0.86 / {\color{white}0}5.13 & 2.01 / 0.86 / 0.88 / {\color{white}0}6.68 & \limebox{1.99} / 0.81 / 0.84 / {\color{white}0}5.13 & \limebox{1.56} / 0.58 / 0.98 / {\color{white}0}5.40 \\
\midrule
\#ReA1
& \pinkbox{87.59} & \pinkbox{86.75} & \pinkbox{73.17} & \pinkbox{95.67}
& \limebox{1.56} / 0.68 / 0.86 / {\color{white}0}5.13 & \limebox{2.01} / 0.86 / 0.88 / {\color{white}0}6.68 & \limebox{1.99} / 0.81 / 0.84 / {\color{white}0}5.13 & \limebox{1.56} / 0.58 / 0.98 / {\color{white}0}5.40 \\
\#ReA3 
& 92.35 & 93.84 & 85.17 & 98.24
& 1.84 / 1.34 / 0.86 / {\color{white}0}9.31 & 2.34 / 1.56 / 0.89 / {\color{white}0}9.47 & 2.58 / 1.63 / 0.87 / 11.81 & 1.69 / 1.04 / 0.99 / {\color{white}0}9.60 \\

\#ReA5
& 93.63 & 95.70 & 87.80 & 98.82
& 2.05 / 2.02 / 0.86 / 14.05 & 2.53 / 2.15 / 0.88 / 13.93 & 2.94 / 2.41 / 0.87 / 14.00 & 1.73 / 1.34 / 0.98 / 15.45 \\
\#ReA10
& 94.59 & 97.80 & 90.94 & 99.38
& 2.49 / 3.57 / 0.86 / 31.32 & 2.81 / 3.31 / 0.89 / 25.35 & 3.63 / 4.15 / 0.87 / 25.02 & 1.81 / 1.93 / 0.98 / 27.07 \\
\#ReA20
& \limebox{95.10} & \limebox{99.03} & \limebox{92.85} & \limebox{99.68}
& \pinkbox{3.12} / 6.18 / 0.86 / 46.17 & \pinkbox{3.09} / 4.62 / 0.88 / 46.13 & \pinkbox{4.62} / 7.23 / 0.86 / 48.38 & \pinkbox{1.89} / 2.72 / 0.99 / 49.27\\
\bottomrule
\end{tabular}
\begin{tablenotes}
    \item ${}^{\text{Note 1~}}$\#ReA -- Number of maximally allowed re-selection attempts.
    \item ${}^{\text{Note 2~}}$The text highlighted with a green background represents the best values, while the text highlighted with a yellow background indicates the worst values. All tables in this paper follow this coloring convention.
\end{tablenotes}
\end{threeparttable}
\end{table*}

\subsubsection{Numerical IK Solver}
The results in Table \ref{tab:ablation2} evaluate the impact of different numerical solvers in finding a solution based on selected seeds. Similar to global settings, the maximum number of allowed numerical iterations per attempt is limited to 7. The DLS method is a numerical IK approach based on damped least squares, while PINV is based on the pseudo-inverse of the Jacobian matrix. Both methods proceed without explicitly handling joint limit violations. If the solvers encounter a joint limit violation during iteration, they simply ignore it and continues to iterate. A validity check is performed only after convergence and before returning the result. Additionally, if a local minimum is encountered during iteration, the solvers reports reports a failure. $\text{PINV}_\text{RR}$ is a revised PINV method. When encountering a local minimum or a limit violation, it randomly re-selects a new seed and restarts the iteration process. Clamping Weighted Least-Norm (CWLN) is a reproduced implementation of the method proposed by \citet{huang2016clamping}. It introduces Jacobian weight adjustments to pull back joints that exceed their motion limits during the iteration process. Clamping Weighted Pseudo Inverse (CWPI) is the pseudo-inverse version of the CWLN method, applying similar weight adjustments to manage boundary constraints while leveraging the computational simplicity of the pseudo-inverse approach. To ensure convergence speed, the DLS and PINV methods are designed to terminate immediately and report failure if the distance between the TCP pose and the target increases in the next iteration compared to the previous one. This constraint helps prevent unnecessary iterations in diverging scenarios, thereby improving computational speed. In contrast, for the CWLN and CWPI methods, which were originally designed to perform corrective adjustments during the iteration process to prevent joint limit violations, this monotonic constraint is not enforced. These methods allow for non-monotonic variations in the distance between the TCP pose and the target pose, as the Jacobian weight adjustments may temporarily increase the distance while ultimately aiming for convergence. 

\begin{figure*}[ht]
    \centering
    \includegraphics[width=\linewidth]{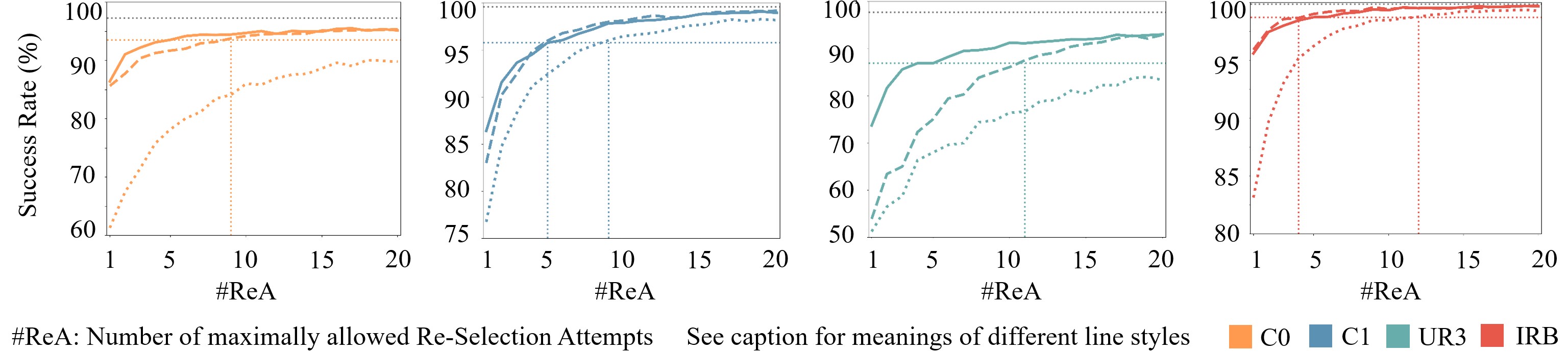}
    \caption{Comparison of different seed selection methods in terms of success rate and average inference time with respect to maximally allowed re-selection attempts. The solid line corresponds to the proposed joint adjustment and re-selection framework, the dashed line corresponds to sorted by only joint space adjustment, and the dotted line corresponds to the workspace proximity-based selection method.}
    \label{seed_sr_avg_t}
\end{figure*}  

\begin{figure*}[ht]
    \centering
    \includegraphics[width=\linewidth]{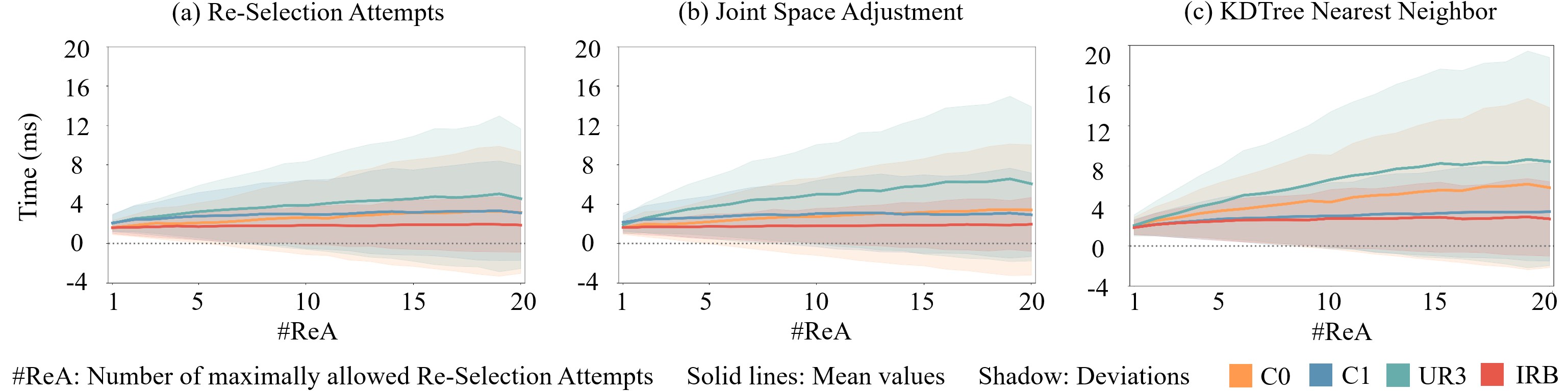}
    \caption{Inference time mean and standard deviation with respect to different maximum re-selection attempts. The shadow regions indicate the standard deviations above and below the mean.}
    \label{seed_std_t}
\end{figure*}  

From the results, we observe that the CWLN and CWPI solvers achieve the highest success rates across all tested robots. This can be attributed to the effectiveness of the weighted joint velocity adjustment and clamping mechanism in preventing joint limit violations. Nevertheless, the differences among the numerical solvers are not particularly pronounced. The success rate variations among the various methods remain around 5\% except for $\text{PINV}_\text{RR}$. The distinct behavior of $\text{PINV}_\text{RR}$ is likely due to its random restart upon encountering joint limit violations, without any buffering or corrective adjustments. As a conclusion, when utilizing the proposed seed selection method, the improvement brought by the numerical solver only yields marginal benefits. The proposed method effectively facilitates rapid convergence within a small number of iterations.

Regarding time efficiency, the CWPI and CWLN have higher time costs compared with other solvers and have the largest maximum inference time across all manipulators. Besides, $\text{PINV}_\text{RR}$ also presents a higher maximum time cost on UR3 and C1. The restart mechanism lets the solver to re-initialize the iteration with a random seed when trapped, thus increasing computation time in certain cases. Considering both the success rate and computational efficiency, we recommend choosing the DLS solver as the default numerical solver to achieve a balanced trade-off.

\subsubsection{Seed Selection Parameters}
Table \ref{tab:ablation3} ablates the seed selection settings. The first and second rows correspond to the results where the first seed is selected based on the workspace nearest neighbor and the proposed method (joint space adjustment plus one time of re-selection attempt), respectively. The results indicated that the proposed method significantly improves the success rate across all testing robots compared with querying the nearest neighbor using the KDTree. The proposed method at least increases the success rate by 10\%, with the highest improvement observed on C0, reaching an increase of up to 26.3\%.

Rows three to seven of Table \ref{tab:ablation3} present the results when the number of re-selection attempts is changed. The numerical suffix in the first column indicates the maximum allowed number of re-selection attempts at each row. It can be observed from the results that with the maximum attempts increasing from 1 to 20, the success rate exhibits steady improvement among all manipulators. However, increasing the maximum re-selection attempts also incurs additional computational overhead, as illustrated by time cost columns. Considering the trade-off of the time cost, increasing the maximum allowed attempts up to a reasonable number (i.e., \#ReA5) can receive a balanced performance on success rate and speed.

\subsubsection{Efficacy of the Proposed Re-Selection Attempts}
Fig. \ref{seed_sr_avg_t} presents more complete success rate comparison results among different seed selection methods. The solid curve corresponds to the proposed joint space adjustment-based selection and re-selection method. The dashed curve corresponds to a seed selection method by solely sorting joint space adjustment values. When a previously selected seed fails, the joint configurations with the next smallest joint space adjustment values are attempted sequentially. The dotted curve corresponds to the workspace proximity-based selection method. The horizontal gray dotted line denotes the near-optimal success rate (exhausting 100 attempts). The horizontal dotted line with the manipulator-specific color indicates the success rate achieved by incorporating the re-selection method with 5 attempts, while the vertical dotted lines with the manipulator-specific color show the attempts of other baselines to reach the approximate success rate. Results suggest that the proposed selection and re-selection method consistently outperforms selection solely based on joint space adjustment. Furthermore, the proposed framework demonstrates superior performance compared to directly using the nearest neighbor in the workspace, especially for the C0 and UR3 manipulators, where exhausting all 20 attempts fails to achieve an approximate success rate. Fig. \ref{seed_std_t} further shows the detailed average time and time deviations with respect to different numbers of maximally allowed re-selection attempts among different seed selection methods. The results indicate that the proposed method largely reduces the IK solving time on C0, UR3, and IRB manipulators, although only a minor advantage is observed on C1. Our observations suggest that the small variation in performance across different selection methods for C1 is primarily due to its large link lengths, which reduce the relative impact of positional differences. In this case, when applying the same weighted distance metric for position and rotation as in C0, the proportional relationship between these components shifts, leading to different performance characteristics. Additionally, the proposed method's time costs present a lower standard deviation across varying attempt numbers. It has a more stable IK-solving performance.

\subsection{Benchmark Comparison with Other Solvers}

We also conducted experiments to compare the proposed method with other numerical-based, analytical-numerical hybrid, and learning-based approaches. For numerical-based and analytical-numerical hybrid approaches, we compared the performance of the proposed method with TRAC-IK \cite{beeson2015trac} and IK-Geo \cite{elias2025ik}. TRAC-IK concurrently integrates the numerical iteration and non-linear optimization to improve the solving efficiency and success rate. IK-Geo decomposes the IK problem for the 6R manipulator into several sub-problems, and each sub-problem can be solved analytically or numerically on a low-dimensional manifold. For learning-based methods, we compared the proposed method with Mixture Density Networks (MDN) \cite{thompson2024modeling}, IKFlow \cite{ames2022ikflow}, and IKDP \cite{tsui2024ikdp}\footnote{IKFlow is implemented using the publicly available code. MDN and IKDP are reproduced following published papers.}. 
In most cases, learning-based IK solvers cannot achieve comparable precision with numerical-based methods (millimeter-level position error vs. $10^{-3}$ millimeter-level position error). Thus, besides directly generating joint configuration, we also evaluate the performance of using the neural network output as an initial seed for numerical IK iteration. The training dataset for the learning-based methods was generated by using forward kinematics to calculate the target positions and rotations given random sampled joint configurations within limits. Quaternion was used for rotation representation. During the inference, the trained neural networks predict the joint configuration given target poses.

\subsubsection{Traditional IK Solvers}
Table \ref{tab:trad_ik} presents a performance comparison against the TRAC-IK method. The results show that TRAC-IK has a relatively low success rate across all robots, ranging from 55\% to 72\%. Meanwhile, it maintains relatively low and stable computation costs, with average time around 2 ms and standard deviation below 1 ms. 

Table \ref{tab:ikgeo} presents the results of IK-Geo. For the implementation of 1D search in the IK-Geo, we discretized the joint range into 36 divisions to detect potential zero-crossing points. Compared to the results in Table \ref{tab:trad_ik}, the success rate of IK-Geo surpasses that of TRAC-IK and the proposed numerical solver, however, the inference time is significantly higher due to the low looping efficiency of our Python implementation. For direct generation, the average position and rotation error remain lower for C0, while the position error in C1 is notable. Moreover, the high standard deviation and max error in both manipulators indicate that the performance of IK-Geo may weaken in certain scenarios given a sparse joint range search space. Utilizing IK-Geo as a seed generator significantly improves the accuracy, ensured by the termination condition in numerical iterations, while maintaining a success rate reduction within 1\%. Although the inference time increased compared with the direct generation, the superior performance is particularly suitable for the most challenging scenarios.

As shown in Table \ref{tab:trad_ik}, the proposed solver, with one permitted re-selection attempt, achieves a high success rate, ranking second to IK-Geo. Notably, it demonstrates particularly strong performance on IRB (96.79\%) and C0 (89.74\%). More importantly, the proposed solver exhibits satisfactory computational efficiency, maintaining an average inference time below 3.5 ms and a standard deviation below 2 ms. Overall, the proposed method effectively balances success rate and computational speed, indicating its potential suitability for time-sensitive applications.

\begin{table*}[ht]
\centering
\caption{Performance Comparison of Traditional IK Solvers}
\label{tab:trad_ik}
\begin{tabular}{l@{\hskip 11pt}c@{\hskip 5.5pt}c@{\hskip 5.5pt}c@{\hskip 5.5pt}c@{\hskip 11pt}c@{\hskip 8.5pt}c@{\hskip 8.5pt}c@{\hskip 8.5pt}c}
\toprule 
 & \multicolumn{4}{c}{Success Rate (\%)} 
 & \multicolumn{4}{c}{Time Cost (Mean / Std / Min / Max Time (ms))} \\ 
\cmidrule(lr){2-5} \cmidrule(lr){6-9}  
& {C0} & {C1} & {UR3} & {IRB} 
& {C0} & {C1} & {UR3} & {IRB}  \\
\midrule
TRAC-IK & 54.48 & 69.95 & 56.37 & 71.85 & 2.07 / 0.61 / 0.86 / {\color{white}0}4.53 & {\color{white}0}2.07 / 0.53 / 0.87 / {\color{white}0}3.16 & 2.09 / 0.57 / 0.86 / {\color{white}0}3.14 & 2.11 / 0.70 / 1.01 / {\color{white}0}5.53 \\
Proposed & 89.74 & 88.18 & 77.93 & 96.79 & 2.17 / 1.54 / 0.93 / 10.49 & {\color{white}0}2.71 / 1.52 / 0.99 / {\color{white}0}9.20 & 3.19 / 1.85 / 0.95 / 16.46 & 1.93 / 1.14 / 1.06 / 13.68 \\

\bottomrule
\end{tabular}
\end{table*}

\begin{table*}[ht]
\centering
\caption{Performance of IK-Geo}
\label{tab:ikgeo}
\begin{threeparttable}
\begin{tabular}{l@{\hskip 8pt}c c@{\hskip 6pt}c@{\hskip 6pt}c@{\hskip 8pt}c c@{\hskip 8pt}c}
\toprule 
& 
\multicolumn{1}{c}{}
& \multicolumn{4}{c}{Direct 1D Search}
& \multicolumn{2}{c}{Output as Seed} \\
\cmidrule(lr){3-6} \cmidrule(lr){7-8}
 & Rbt 
 & {Time} & {Succ. Rate (\%)} & {\begin{tabular}[c]{@{}c@{}}Position Error (mm)\\ (Mean / Std / Min / Q1 / Q3 / Max)\end{tabular}} & {\begin{tabular}[c]{@{}c@{}}Rotation Error ($^{\circ}$)\\ (Mean / Std / Min / Q1 / Q3 / Max)\end{tabular}} 
 & {Time} & {Succ. Rate (\%)} \\

\midrule
\multirow{2}{*}{IK-Geo} 
& C0 & {\color{white}0}8.70 & 94.06 & {\color{white}0}6.96 / {\color{white}0}8.99 / 0.00 / {\color{white}0}2.89 / {\color{white}0}8.93 / 133.56 & 1.67 / 5.93 / 0.00 / 0.12 / 1.13 / 88.84 & {\color{white}0}9.64 & 93.61 \\
& C1 & 15.33 & 98.54 & 31.56 / 46.80 / 0.00 / 11.55 / 35.38 / 495.95 & 2.90 / 8.14 / 0.00 / 0.18 / 2.04 / 87.46 & 21.63 & 98.47 \\
\bottomrule
\end{tabular}
\end{threeparttable}
\begin{tablenotes}
    \item ${}^{\text{Note~}}$ For both robots, we perform the search for zero-crossing points within the 1D space of the fourth joint. To accelerate the process, we discretize the joint range into 36 divisions and identify the division where the sign of the error changes. Then, we apply linear interpolation between the two endpoints of that division to generate a candidate solution. In practice, the interpolation can be repeated using binary subdivision to improve accuracy, provided that practitioners do not mind the increased computation time. However, in our experiments, we did not adopt this refinement step.
\end{tablenotes}
\end{table*}

\subsubsection{Learning-based IK Solvers}

\begin{table*}[ht]
\centering
\caption{Performance Comparison of learning-based Methods on 2,000 Random IK Problems}
\label{tab:learning}
\begin{threeparttable}
\begin{tabular}{l l c ccccc ccccc}
\toprule 
& 
\multicolumn{1}{c}{}
& \multicolumn{3}{c}{Direct Inference / Direct Generation} 
& \multicolumn{2}{c}{First as Seed} 
& \multicolumn{2}{c}{Best as Seed} \\
\cmidrule(lr){3-5} \cmidrule(lr){6-7} \cmidrule(lr){8-9}
 & Robot 
 & {Time} & {\begin{tabular}[c]{@{}c@{}}Position Error (mm)\\ (Mean / Min / Max / Q1 / Q3)\end{tabular}} & {\begin{tabular}[c]{@{}c@{}}Rotation Error ($^{\circ}$)\\ (Mean / Min / Max / Q1 / Q3)\end{tabular}} 
 & {Time} & {\begin{tabular}[c]{@{}c@{}}Succ\\(\%)\end{tabular}} & {Time} & {\begin{tabular}[c]{@{}c@{}}Succ\\(\%)\end{tabular}}\\

\midrule
\multirow{4}{*}{MDN} 
& C0   & 0.39 & {\color{white}0}17.77 / ~---~ / ~---~ / ~---~ / ~~--- & ~~7.79 / ~---~ / ~---~ / ~---~ / ~~--- & 1.60 & 87.80 & --- & --- \\
& C1 & 0.43 & 141.89 / ~---~ / ~---~ / ~---~ / ~~--- & ~87.15 / ~---~ / ~---~ / ~---~ / ~~--- & 2.52 & 59.30 & --- & --- \\
& UR3      & 0.42 & 140.67 / ~---~ / ~---~ / ~---~ / ~~--- & 118.70 / ~---~ / ~---~ / ~---~ / ~~--- & 3.05 & 25.35 & --- & --- \\
& IRB & 0.40 & {\color{white}0}68.04 / ~---~ / ~---~ / ~---~ / ~~--- & ~19.14 / ~---~ / ~---~ / ~---~ / ~~---  & 1.70 & 92.90 & --- & --- \\
\midrule
\multirow{4}{*}{IKFlow} 
& C0   & 3.74 & {\color{white}0}7.58 / {\color{white}0}1.18 / {\color{white}0}40.51 / {\color{white}0}3.15 / {\color{white}00}8.57 & {\color{white}00}4.65 / {\color{white}0}0.77 / {\color{white}0}25.89 / {\color{white}0}2.01 / {\color{white}00}5.11 & 4.65 & 88.98 & 14.06 & 96.72 \\
& C1 & 4.53 & 50.67 / {\color{white}0}6.51 / 355.34 / 18.42 / {\color{white}0}51.90 & {\color{white}0}31.48 / {\color{white}0}1.74 / 143.01 / {\color{white}0}7.27 / {\color{white}0}41.41 & 5.12 & 82.51 & 14.13 & 95.10 \\
& UR3      & 4.06 & 90.89 / 10.43 / 393.33 / 37.70 / 113.38 & 101.91 / 15.86 / 175.70 / 63.69 / 141.90 & 7.23 & 42.61 & 16.27 & 88.46 \\
& IRB & 4.04 & 26.13 / {\color{white}0}2.98 / 155.21 / {\color{white}0}8.63 / {\color{white}0}28.41 & {\color{white}00}7.98 / {\color{white}0}0.94 / {\color{white}0}54.84 / {\color{white}0}2.62 / {\color{white}00}8.07 & 4.61 & 90.66 & 14.05 & 98.56 \\
\midrule
\multirow{4}{*}{IKDP} 
& C0   & {\color{white}0}7.72 & {\color{white}0}8.77 / 4.22 / {\color{white}0}18.66 / 6.73 / 10.26 & {\color{white}0}3.63 / 1.71 / {\color{white}00}8.65 / 2.77 / {\color{white}0}4.17 & 9.25 & 94.22 & 15.65 & 95.95 \\
& C1 & 15.77 & 18.74 / 3.95 / {\color{white}0}97.11 / 9.32 / 20.92 & {\color{white}0}6.74 / 0.72 / {\color{white}0}62.81 / 1.83 / {\color{white}0}5.58 & 16.34 & 90.06 & 24.44 & 96.50 \\
& UR3      & {\color{white}0}8.12 & 28.49 / 2.65 / 251.16 / 7.38 / 23.82 & 25.52 / 1.40 / 143.93 / 4.58 / 28.84 & 9.78 & 80.70 & 16.32 & 94.62 \\
& IRB & {\color{white}0}7.88 & 21.88 / 2.57 / 149.21 / 7.41 / 21.88 &{\color{white}0}7.35 / 0.68 / {\color{white}0}67.79 / 1.87 / {\color{white}0}6.07 & 10.23 & 90.81 & 17.16 & 97.59 \\
\bottomrule
\end{tabular}
\end{threeparttable}
\begin{tablenotes}
    \item $^{\text{Note 1}}$ Each model was trained separately for the four robots.
    \item $^{\text{Note 2}}$ The listed descriptive statistics in this table represent the average of each batch solution's statistics across 2,000 random IK problems.
    \item $^{\text{Note 3}}$ \parbox[t]{.95\linewidth}{For IKFlow and IKDP, a batch of 32 joint value solutions is sampled per IK problem. During the inference of MDN, the Gaussian component with the highest probability is selected, leading to a deterministic joint configuration.}
    \item $^{\text{Note 4}}$ \parbox[t]{.95\linewidth}{The last four columns present the performance of using the model's output as the seed joint values for numerical IK iterations. Due to the diversity of solutions from generative models, we separately examined the performance of the first generated solution and the best solution among the solution batch (the one least distant from the target end-effector pose).}
\end{tablenotes}
\end{table*} 

Table \ref{tab:learning} presents the results of learning-based IK solvers. The table includes two main sections. For direct inference, MDN is almost unable to generate approximate joint configurations, except for the C0 robot (only finite possible joint configurations for the desired target). Although MDN can model the probability distribution of the dataset through a Gaussian mixture model, it remains insufficient for complex IK problems. IKFlow significantly decreases the position and rotation error compared with MDN, especially reaching the highest accuracy for the C0 robot among the three models. However, the model's performance appears to be sensitive to network parameters. Current experiment results are derived from a uniform architecture of 15 coupling layers and a latent space dimension of 15, yet the results vary among different robots. This suggests that more careful parameter selection is required to exploit the potential of the IKFlow model fully. Furthermore, the IKDP method exhibits the highest position and rotation precision compared with IKFlow and MDN on most robots, demonstrating its ability in complex solution space modeling. However, it needs more inference time to conduct iterative noise denoising and generate the solution, which limits its application in time-sensitive scenarios. Overall, the solutions generated by these learning-based methods are quite diverse while closely approximating the desired target.

We also evaluated the performance of using these generated solutions as initial seeds for numerical iterations. The results are shown in the last two columns of Table \ref{tab:learning}. Due to the large variation within each batch of solutions, we separately evaluate using the first generated sample (prioritize speed) and selecting the best solution within the batch (prioritize accuracy). The results indicate that the success rate of using the generated solution as seeds is highly dependent on the generation quality. For example, due to the poor performance of the MDN and IKFlow model on the UR3 robot, their success rates are 25.35\% and 42.61\%, respectively. Moreover, selecting the best solution from the batch consistently outperforms using the first solution in success rate. For example, a large increase can be observed in the UR3 manipulator where the success rate improved 45\% in the IKFlow-based method. Notably, utilizing the best solution as an initial seed leads to a high success rate compared to our proposed method across all robots. The result highlights the generalization ability and great potential of generative models in solving IK problems.

For the inference time, the MDN model outperforms other models with an average inference time below 0.5 ms. IKFlow and IKDP solvers require 4$\sim$5 ms and 7$\sim$15 ms for direct inference, respectively\footnote{For IKDP, we set the minimum inference steps while maintaining sample quality: 5 steps for IRB, C0, and UR3, and 10 steps for C1.}. Such results suggest that the generative models are more suitable for scenarios requiring solution diversity rather than real-time performance.

\section{Conclusions and Future Work} 
We developed a novel seed selection method for fast numerical IK solutions. The selection strategy involved first narrowing down seed candidates based on workspace proximity, followed by sorting the candidates according to joint space adjustment. Crucially, we introduced a re-selection attempt mechanism, which allowed the solver to circumvent local minima and joint limit constraints after failed iterations. Experimental results demonstrated the effectiveness of the proposed method, showing a balanced performance in terms of success rate and computational efficiency when compared with both conventional and learning-based IK solvers. The method is well-suited for time-sensitive applications. 

Since our method is essentially a numerical one, it can naturally incorporate task constraints into the solving process, similar to traditional numerical solvers. The incorporation enables the manipulator to utilize specific joints to accomplish secondary tasks during IK solving while still maintaining the balanced performance characteristics offered by the proposed seed selection strategy. We are interested in exploring this direction further in future work.

\section*{Appendix}

The following table shows the sample number and the corresponding KDTree query file sizes of Small, Medium, and Large scales compared in Table \ref{tab:ablation1}.

\begin{table}[ht]
\centering
\caption{Details of the KDTree sizes in Table \ref{tab:ablation1}}
\label{tab:kdtree-sizes}
\begin{tabular}{l c@{\hskip 8pt}c c@{\hskip 8pt}c}
\toprule
 & \multicolumn{2}{c}{C0 / C1 / UR3} & \multicolumn{2}{c}{IRB}\\
\cmidrule(lr){2-3}\cmidrule(lr){4-5}
 & \# Samples & \text{Query File Size} 
 & \# Samples & \text{Query File Size} \\
\midrule
Small  
 & {\color{white}0}12{,}000 & {\color{white}0}0.8 MB  
 & {\color{white}0}48{,}000 & {\color{white}0}3.3 MB \\
Medium
 & {\color{white}0}40{,}320 & {\color{white}0}2.8 MB   
 & 201{,}600 & 13.6 MB \\
Large 
 & 151{,}200 & 10.8 MB  
 & 604{,}800 & 43.3 MB \\
\bottomrule
\end{tabular}
\end{table}




\normalem
\bibliographystyle{IEEEtranN}
\bibliography{citations.bib}

\begin{thebibliography}{48}
\providecommand{\natexlab}[1]{#1}
\providecommand{\url}[1]{#1}
\csname url@samestyle\endcsname
\providecommand{\newblock}{\relax}
\providecommand{\bibinfo}[2]{#2}
\providecommand{\BIBentrySTDinterwordspacing}{\spaceskip=0pt\relax}
\providecommand{\BIBentryALTinterwordstretchfactor}{4}
\providecommand{\BIBentryALTinterwordspacing}{\spaceskip=\fontdimen2\font plus
\BIBentryALTinterwordstretchfactor\fontdimen3\font minus \fontdimen4\font\relax}
\providecommand{\BIBforeignlanguage}[2]{{%
\expandafter\ifx\csname l@#1\endcsname\relax
\typeout{** WARNING: IEEEtranN.bst: No hyphenation pattern has been}%
\typeout{** loaded for the language `#1'. Using the pattern for}%
\typeout{** the default language instead.}%
\else
\language=\csname l@#1\endcsname
\fi
#2}}
\providecommand{\BIBdecl}{\relax}
\BIBdecl

\bibitem[Lee and Liang(1988)]{lee1988new}
H.-Y. Lee and C.-G. Liang, ``A new vector theory for the analysis of spatial mechanisms,'' \emph{Mechanism and Machine Theory}, vol.~23, no.~3, pp. 209--217, 1988.

\bibitem[Chiaverini et~al.(1994)Chiaverini, Siciliano, and Egeland]{chiaverini1994review}
S.~Chiaverini, B.~Siciliano, and O.~Egeland, ``Review of the damped least-squares inverse kinematics with experiments on an industrial robot manipulator,'' \emph{IEEE Transactions on Control Systems Technology}, vol.~2, no.~2, pp. 123--134, 1994.

\bibitem[Pieper(1969)]{pieper1969kinematics}
D.~L. Pieper, \emph{The kinematics of manipulators under computer control}.\hskip 1em plus 0.5em minus 0.4em\relax Stanford University, 1969.

\bibitem[Tsai and Morgan(1985)]{tsai1985solving}
L.-W. Tsai and A.~P. Morgan, ``Solving the kinematics of the most general six-and five-degree-of-freedom manipulators by continuation methods,'' \emph{Journal of Mechanisms, Transmissions, and Automation in Design}, vol. 107, no.~2, 1985.

\bibitem[Raghavan and Roth(1990)]{raghavan1990kinematic}
M.~Raghavan and B.~Roth, ``Kinematic analysis of the 6r manipulator of general geometry,'' in \emph{International Symposium on Robotics Research}, 1990, pp. 314--320.

\bibitem[Raghavan and Roth(1993)]{raghavan1993inverse}
------, ``Inverse kinematics of the general 6r manipulator and related linkages,'' \emph{Journal of Mechanisms, Transmissions, and Automation in Design}, vol. 115, no.~3, 1993.

\bibitem[Lee et~al.(1991)Lee, Woernle, and Hiller]{lee1991complete}
H.~Lee, C.~Woernle, and M.~Hiller, ``A complete solution for the inverse kinematic problem of the general 6r robot manipulator,'' \emph{Journal of Mechanisms, Transmissions, and Automation in Design}, vol. 113, no.~4, 1991.

\bibitem[Kohli and Osvatic(1993)]{kohli1993inverse}
D.~Kohli and M.~Osvatic, ``Inverse kinematics of general 6r and 5r, p serial manipulators,'' \emph{Journal of Mechanisms, Transmissions, and Automation in Design}, vol. 115, no.~4, 1993.

\bibitem[Manocha and Canny(1994)]{manocha1994efficient}
D.~Manocha and J.~F. Canny, ``Efficient inverse kinematics for general 6r manipulators,'' \emph{IEEE Transactions on Robotics and Automation}, vol.~10, no.~5, pp. 648--657, 1994.

\bibitem[Rudny(2014)]{rudny2014solving}
T.~Rudny, ``Solving inverse kinematics by fully automated planar curves intersecting,'' \emph{Mechanism and Machine Theory}, vol.~74, pp. 310--318, 2014.

\bibitem[Diankov(2010)]{diankov2010automated}
R.~Diankov, ``Automated construction of robotic manipulation programs,'' Ph.D. dissertation, Carnegie Mellon University, USA, 2010.

\bibitem[Zhang and Hannaford(2019)]{zhang2017ikbt}
D.~Zhang and B.~Hannaford, ``Ikbt: solving symbolic inverse kinematics with behavior tree,'' \emph{Journal of Aritificial Intelligence Research}, vol.~65, no.~1, p. 457–486, 2019.

\bibitem[Aristidou et~al.(2018)Aristidou, Lasenby, Chrysanthou, and Shamir]{aristidou2018inverse}
A.~Aristidou, J.~Lasenby, Y.~Chrysanthou, and A.~Shamir, ``Inverse kinematics techniques in computer graphics: A survey,'' in \emph{Computer Graphics Forum}, vol.~37, no.~6, 2018, pp. 35--58.

\bibitem[Elias and Wen(2025)]{elias2025ik}
A.~J. Elias and J.~T. Wen, ``Ik-geo: Unified robot inverse kinematics using subproblem decomposition,'' \emph{Mechanism and Machine Theory}, vol. 209, p. 105971, 2025.

\bibitem[Whitney(1969)]{whitney1969resolved}
D.~E. Whitney, ``Resolved motion rate control of manipulators and human prostheses,'' \emph{IEEE Transactions on Man-Machine Systems}, vol.~10, no.~2, pp. 47--53, 1969.

\bibitem[Orin and Schrader(1984)]{orin1984efficient}
D.~E. Orin and W.~W. Schrader, ``Efficient computation of the jacobian for robot manipulators,'' \emph{International Journal of Robotics Research}, vol.~3, no.~4, pp. 66--75, 1984.

\bibitem[Nakamura and Hanafusa(1986)]{nakamura1986inverse}
Y.~Nakamura and H.~Hanafusa, ``Inverse kinematic solutions with singularity robustness for robot manipulator control,'' \emph{Journal of Dynamic Systems, Measurement and Control}, vol. 108, pp. 163--171, 1986.

\bibitem[Wampler(1986)]{wampler1986manipulator}
C.~W. Wampler, ``Manipulator inverse kinematic solutions based on vector formulations and damped least-squares methods,'' \emph{IEEE Transactions on Systems, Man, and Cybernetics}, vol.~16, no.~1, pp. 93--101, 1986.

\bibitem[Sugihara(2011)]{sugihara2011solvability}
T.~Sugihara, ``Solvability-unconcerned inverse kinematics by the levenberg--marquardt method,'' \emph{IEEE Transactions on Robotics}, vol.~27, no.~5, pp. 984--991, 2011.

\bibitem[Mayorga et~al.(1992)Mayorga, Wong, and Milano]{mayorga1992fast}
R.~V. Mayorga, A.~K. Wong, and N.~Milano, ``A fast procedure for manipulator inverse kinematics evaluation and pseudoinverse robustness,'' \emph{IEEE Transactions on Systems, Man, and Cybernetics}, vol.~22, no.~4, pp. 790--798, 1992.

\bibitem[Lenar{\v{c}}i{\v{c}}(1985)]{lenarvcivc1985efficient}
J.~Lenar{\v{c}}i{\v{c}}, ``An efficient numerical approach for calculating the inverse kinematics for robot manipulators,'' \emph{Robotica}, vol.~3, no.~1, pp. 21--26, 1985.

\bibitem[Zhao and Badler(1994)]{zhao1994inverse}
J.~Zhao and N.~I. Badler, ``Inverse kinematics positioning using nonlinear programming for highly articulated figures,'' \emph{ACM Transactions on Graphics}, vol.~13, no.~4, pp. 313--336, 1994.

\bibitem[Sekiguchi and Takesue(2021)]{sekiguchi2021numerical}
M.~Sekiguchi and N.~Takesue, ``Numerical method for inverse kinematics using an extended angle-axis vector to avoid deadlock caused by joint limits,'' \emph{Advanced Robotics}, vol.~35, no.~15, pp. 919--926, 2021.

\bibitem[Huang et~al.(2016)Huang, Peng, Wei, and Xiang]{huang2016clamping}
S.~Huang, Y.~Peng, W.~Wei, and J.~Xiang, ``Clamping weighted least-norm method for the manipulator kinematic control with constraints,'' \emph{International Journal of Control}, vol.~89, no.~11, pp. 2240--2249, 2016.

\bibitem[Farzan and DeSouza(2013)]{farzan2013dh}
S.~Farzan and G.~N. DeSouza, ``From dh to inverse kinematics: A fast numerical solution for general robotic manipulators using parallel processing,'' in \emph{IEEE/RSJ International Conference on Intelligent Robots and Systems (IROS)}, 2013, pp. 2507--2513.

\bibitem[Demby’s et~al.(2019)Demby’s, Gao, and DeSouza]{demby2019study}
J.~Demby’s, Y.~Gao, and G.~N. DeSouza, ``A study on solving the inverse kinematics of serial robots using artificial neural network and fuzzy neural network,'' in \emph{IEEE International Conference on Fuzzy Systems (FUZZ-IEEE)}, 2019, pp. 1--6.

\bibitem[Csiszar et~al.(2017)Csiszar, Eilers, and Verl]{csiszar2017solving}
A.~Csiszar, J.~Eilers, and A.~Verl, ``On solving the inverse kinematics problem using neural networks,'' in \emph{International Conference on Mechatronics and Machine Vision in Practice (M2VIP)}, 2017, pp. 1--6.

\bibitem[Xia and Wang(2001)]{xia2001dual}
Y.~Xia and J.~Wang, ``A dual neural network for kinematic control of redundant robot manipulators,'' \emph{IEEE Transactions on Systems, Man, and Cybernetics, Part B (Cybernetics)}, vol.~31, no.~1, pp. 147--154, 2001.

\bibitem[Oyama et~al.(2001)Oyama, Chong, Agah, and Maeda]{oyama2001inverse}
E.~Oyama, N.~Y. Chong, A.~Agah, and T.~Maeda, ``Inverse kinematics learning by modular architecture neural networks with performance prediction networks,'' in \emph{IEEE International Conference on Robotics and Automation (ICRA)}, vol.~1, 2001, pp. 1006--1012.

\bibitem[Lendaris et~al.(1999)Lendaris, Mathia, and Saeks]{lendaris1999linear}
G.~G. Lendaris, K.~Mathia, and R.~Saeks, ``Linear hopfield networks and constrained optimization,'' \emph{IEEE Transactions on Systems, Man, and Cybernetics, Part B (Cybernetics)}, vol.~29, no.~1, pp. 114--118, 1999.

\bibitem[D'Souza et~al.(2001)D'Souza, Vijayakumar, and Schaal]{d2001learning}
A.~D'Souza, S.~Vijayakumar, and S.~Schaal, ``Learning inverse kinematics,'' in \emph{IEEE/RSJ International Conference on Intelligent Robots and Systems (IROS)}, vol.~1, 2001, pp. 298--303.

\bibitem[Phaniteja et~al.(2017)Phaniteja, Dewangan, Guhan, Sarkar, and Krishna]{phaniteja2017deep}
S.~Phaniteja, P.~Dewangan, P.~Guhan, A.~Sarkar, and K.~M. Krishna, ``A deep reinforcement learning approach for dynamically stable inverse kinematics of humanoid robots,'' in \emph{IEEE International Conference on Robotics and Biomimetics (ROBIO)}.\hskip 1em plus 0.5em minus 0.4em\relax IEEE, 2017, pp. 1818--1823.

\bibitem[Guo et~al.(2019)Guo, Huang, Ren, and Wang]{guo2019reinforcement}
Z.~Guo, J.~Huang, W.~Ren, and C.~Wang, ``A reinforcement learning approach for inverse kinematics of arm robot,'' in \emph{International Conference on Robotics, Control and Automation}, 2019, pp. 95--99.

\bibitem[Malik et~al.(2022)Malik, Lischuk, Henderson, and Prazenica]{malik2022deep}
A.~Malik, Y.~Lischuk, T.~Henderson, and R.~Prazenica, ``A deep reinforcement-learning approach for inverse kinematics solution of a high degree of freedom robotic manipulator,'' \emph{Robotics}, vol.~11, no.~2, p.~44, 2022.

\bibitem[Perrusqu{\'\i}a et~al.(2021)Perrusqu{\'\i}a, Yu, and Li]{perrusquia2021multi}
A.~Perrusqu{\'\i}a, W.~Yu, and X.~Li, ``Multi-agent reinforcement learning for redundant robot control in task-space,'' \emph{International Journal of Machine Learning and Cybernetics}, vol.~12, pp. 231--241, 2021.

\bibitem[Zhang(2022)]{zhang2022simulation}
Z.~Zhang, ``Simulation of robotic arm grasping control based on proximal policy optimization algorithm,'' in \emph{Journal of Physics: Conference Series}, vol. 2203, no.~1, 2022, p. 012065.

\bibitem[Zhao et~al.(2024)Zhao, Wei, Xiao, Sun, Zhang, Guo, and Yang]{zhao2024inverse}
C.~Zhao, Y.~Wei, J.~Xiao, Y.~Sun, D.~Zhang, Q.~Guo, and J.~Yang, ``Inverse kinematics solution and control method of 6-degree-of-freedom manipulator based on deep reinforcement learning,'' \emph{Scientific Reports}, vol.~14, no.~1, p. 12467, 2024.

\bibitem[Ardizzone et~al.(2019)Ardizzone, Kruse, Rother, and Köthe]{ardizzone2018analyzing}
L.~Ardizzone, J.~Kruse, C.~Rother, and U.~Köthe, ``Analyzing inverse problems with invertible neural networks,'' in \emph{International Conference on Learning Representations (ICLR)}, 2019.

\bibitem[Kruse et~al.(2021)Kruse, Ardizzone, Rother, and K{\"o}the]{kruse2021benchmarking}
J.~Kruse, L.~Ardizzone, C.~Rother, and U.~K{\"o}the, ``Benchmarking invertible architectures on inverse problems,'' \emph{arXiv preprint arXiv:2101.10763}, 2021.

\bibitem[Ames et~al.(2022)Ames, Morgan, and Konidaris]{ames2022ikflow}
B.~Ames, J.~Morgan, and G.~Konidaris, ``Ikflow: Generating diverse inverse kinematics solutions,'' \emph{IEEE Robotics and Automation Letters}, vol.~7, no.~3, pp. 7177--7184, 2022.

\bibitem[Ren and Ben-Tzvi(2020)]{ren2020learning}
H.~Ren and P.~Ben-Tzvi, ``Learning inverse kinematics and dynamics of a robotic manipulator using generative adversarial networks,'' \emph{Robotics and Autonomous Systems}, vol. 124, p. 103386, 2020.

\bibitem[Ho et~al.(2023)Ho, Chan, King, and Yen]{ho2023deep}
C.-K. Ho, L.-W. Chan, C.-T. King, and T.-Y. Yen, ``A deep learning approach to navigating the joint solution space of redundant inverse kinematics and its applications to numerical ik computations,'' \emph{IEEE Access}, vol.~11, pp. 2274--2290, 2023.

\bibitem[Limoyo et~al.(2024)Limoyo, Mari{\'c}, Giamou, Alexson, Petrovi{\'c}, and Kelly]{limoyo2024generative}
O.~Limoyo, F.~Mari{\'c}, M.~Giamou, P.~Alexson, I.~Petrovi{\'c}, and J.~Kelly, ``Generative graphical inverse kinematics,'' \emph{IEEE Transactions on Robotics}, 2024.

\bibitem[Beeson and Ames(2015)]{beeson2015trac}
P.~Beeson and B.~Ames, ``Trac-ik: An open-source library for improved solving of generic inverse kinematics,'' in \emph{IEEE-RAS International Conference on Humanoid Robots (Humanoids)}, 2015, pp. 928--935.

\bibitem[Ito et~al.(2017)Ito, Kawatsu, and Shibata]{ito2017kinematic}
M.~Ito, K.~Kawatsu, and M.~Shibata, ``Kinematic control of redundant manipulators for admitting joint range of motion maximally,'' \emph{IEEJ Journal of Industry Applications}, vol.~6, no.~4, pp. 278--285, 2017.

\bibitem[Nakamura et~al.(1987)Nakamura, Hanafusa, and Yoshikawa]{nakamura1987task}
Y.~Nakamura, H.~Hanafusa, and T.~Yoshikawa, ``Task-priority based redundancy control of robot manipulators,'' \emph{International Journal of Robotics Research}, vol.~6, no.~2, pp. 3--15, 1987.

\bibitem[Thompson et~al.(2024)Thompson, Cho, Brown, and Kuntz]{thompson2024modeling}
J.~Thompson, B.~Y. Cho, D.~S. Brown, and A.~Kuntz, ``Modeling kinematic uncertainty of tendon-driven continuum robots via mixture density networks,'' in \emph{International Symposium on Medical Robotics (ISMR)}, 2024, pp. 1--7.

\bibitem[Tsui et~al.(2024)Tsui, Tuan, and Shuai]{tsui2024ikdp}
H.-T. Tsui, Y.-R. Tuan, and H.-H. Shuai, ``Ikdp: Inverse kinematics through diffusion process,'' \emph{arXiv preprint arXiv:2410.15341}, 2024.

\end{thebibliography}

\end{document}